\documentclass[11pt,a4]{article}

\usepackage{amsmath, hyperref}

\usepackage{type1cm}        
\usepackage{makeidx}         
\usepackage{graphicx}        
\usepackage{multicol}        
\usepackage[bottom]{footmisc}

\usepackage{newtxtext}       %
\usepackage{newtxmath}       

\usepackage{bm}

\usepackage{listings}

\usepackage{caption}
\DeclareCaptionFont{white}{\color{white}}
\DeclareCaptionFormat{listing}{\colorbox{gray}{\parbox{\textwidth}{#1#2#3}}}
\newtheorem{definition}{Definition}
\newtheorem{theorem}{Theorem}

\newcommand{\N}{ \mathbb{N} }

\newcommand{\R}{ \mathbb{R} }

\newcommand{\wh}[1]{ \widehat{ #1 } }
\newcommand{\wt}[1]{ \widetilde{ #1 } }

\newcommand{\calC}{\mathcal{C}}

\newcommand{\calF}{\mathcal{F}}

\newcommand{\eins}{{\bm 1}}

\newcommand{\matX}{{\bm X}}
\newcommand{\matY}{{\bm Y}}

\newcommand{\matW}{{\bm W}}
\newcommand{\matZ}{{\bm Z}}
\newcommand{\vecnull}{{\bm 0}}

\newcommand{\vecb}{{\bm b}}

\newcommand{\vecg}{{\bm g}}

\newcommand{\veco}{{\bm o}}

\newcommand{\vecu}{{\bm u}}

\newcommand{\vecw}{{\bm w}}
\newcommand{\vecx}{{\bm x}}

\newcommand{\vecy}{{\bm y}}
\newcommand{\vecY}{{\bm Y}}
\newcommand{\vecZ}{{\bm Z}}
\newcommand{\vecz}{{\bm z}}

\newcommand{\bfbeta}{\bm\beta}
\newcommand{\bftheta}{{\bm\theta}}

\newcommand{\bfeta}{\bm\eta}

\newcommand{\bfeps}{\bm \epsilon}
\newcommand{\bhbeta}{\widehat{\bm\beta}}

\newcommand{\EE}{\mathbb E}
\newcommand{\PP}{\mathbb P}

\newcommand{\matid}{I}

\makeindex             

\begin{document}

\begin{center}
{\Large \bf Cross-Validation and Uncertainty Determination for Randomized Neural Networks with Applications to Mobile Sensors}

\vskip 0.4cm
{\large Ansgar Steland and Bart E. Pieters}

\vskip 0.4cm
{Institute of Statistics, RWTH Aachen University, W\"ullnerstr. 3, D-52056 Aachen, \href{}{steland@stochastik.rwth-aachen.de} \\ and \\
	Forschungszentrum J\"ulich, Institut f\"ur Energie- und Klimaforschung, IEK-5, D-52425 J\"ulich, \href{}{b.pieters@fz-juelich.de}}

\end{center}

\textbf{Abstract:} 
	Randomized artificial neural networks such as extreme learning machines provide an attractive and efficient method for supervised learning under limited computing ressources and green machine learning. This especially applies when equipping mobile devices (sensors) with weak artificial intelligence.  Results are discussed about supervised learning with such networks and regression methods in terms of consistency and bounds for the generalization and prediction error. Especially, some recent results are reviewed addressing learning with data sampled by moving sensors leading to non-stationary and dependent samples. 
	As randomized networks lead to random out-of-sample performance measures, we study a cross-validation approach to handle the randomness and make use of it to improve out-of-sample performance. Additionally, a computationally efficient approach to determine the resulting uncertainty in terms of a confidence interval for the mean out-of-sample prediction error is discussed based on two-stage estimation. The approach is applied to a prediction problem arising in vehicle integrated photovoltaics.

\section{Introduction}
\label{sec:1}

Artificial neural networks are an attractive class of models for supervised learning tasks arising in data science such as nonlinear regression and predictive analytics. There is a growing interest which is mainly driven by the development of highly efficient and fast algorithms for training and an improved understanding of multilayer deep neural network architectures, especially in terms of how to design and fit them for concrete machine learning tasks. Supervised learning tasks are pervasive and allow to equip devices and technical systems with weak artificial intelligence by processing data collected by sensors. This quickly results in big data sets difficult to handle by, for example, cars or smartphones, which have limited computing ressources but can highly benefit form autonomous learning abilities. For example, one can attach solar panels to cars and trucks and use past and current data as well as planned routes and geographical data to predict the energy production of the solar panels and optimize their usage or storage during driving. 

 Extreme learning machines, \cite{Huang2004}, are widely used in applications due to their extremely fast learning compared to full optimization of feedforward neural networks. For this reason, they have been chosen as a benchmark classifier in the recently updated MNIST data set of handwritten characters, see \cite{eMNIST2017} for details. They also performed very well in empirical comparison studies investigating 179 different classifiers for a large number of publicly available data sets,  \cite{Delgao2014}. Extreme learning machines select the parameters of all layers except the last (output) layer randomly. The weights of the last layer are optimized by minimizing a (regularized) least squares error criterion. Since the output layer uses a linear activation function, this step means that the random but data-dependent features generated by the preceeding layers are linearly combined to explain the target values (responses). The optimization of the parameters of the output layer collapses to a linear least squares problem, which can be solved explicitly and does not require iterative minimization algorithms. Consequently, extreme learning machines optimize only a part of the parameters and choose the remaining ones randomly. Therefore, they belong to the class of randomized networks, see \cite{SW2017} for a broader review including random kernel machines using random Fourier features and reservoid computing based on randomized recurrent networks.
  
In this paper, cross-validation and two-stage estimation methodologies  are proposed to handle the uncertainty resulting from the randomization of feedforward neural networks in a statistical sound way. The basic idea studied here is to apply a simple cross-validation scheme to evaluate network realizations to pick one with good out-of-sample generalization. The approach can be easily combined with model selection, especially the choice of the number of hidden neurons. Since the usual error criteria minimized by training algorithms are known to have many local minima, the random choice of the starting point used for optimization also leads to some degree of randomness of the trained network. This especially applies when using early stopping techniques to achieve better generalization abilities. Hence, the approach can also be used when optimizing all parameters of a neural net. Further, the basic idea can easily be applied to other machine learners, but in our presentation we focus on (randomized) feedforward networks resp. extreme learning machines.

Cross-validation is a well established and widely used statistical approach, \cite{Stone1974},
and has been extensively studied for nonlinear regression and prediction, see, e.g., \cite{Ardot2010} for a review and \cite{Steland2012} for results addressing kernel smoothers for dependent data streams with possible changes. In its simplest form, one splits the data in a training (or learning) sample and a validation sample. The performance of a method estimated (trained) from the learning sample is then evaluated by applying it to the validation sample and tuned by choosing hyperparameters of the method. The final fit is evaluated using a test sample. In principle, there are well established results when this works. Cross-validation for comparing regression procedures has been studied by \cite{Yang2007} for i.i.d. training samples. It is, however, worth mentioning that these results are usually not applicable to machine learning methods such as artificial neural networks, since this would require to train and apply the methods in such a way that consistency as statistical estimators is ensured. But this if frequently not met in practice. 

As a second statistical tool, we discuss the construction of an uncertainty interval for the mean sample prediction error in the validation data set. Two-stage estimation is a well established statistical approach leading to small required sample sizes, thus keeping the computational costs at a low level. 

As a challenging data science problem we discuss the application to a prediction problem arising in photovoltaics (PV) and analyze data from a pilot study to illustrate the proposal. In recent years, especially with the electrification of transport, 
there is an increasing interest in Vehicle Integrated Photovoltaics 
(VIPV) applications \cite{Kuehnel2017VIPV,Birnie2016VIPV}. In this 
context there is a keen interest in photovoltaic (PV) yield prediction for such 
applications. Yield prediction is complicated by the ever changing 
orientation and location of the vehicle, with influences of buildings 
and other objects. Many vehicles have specific routes that often 
repeat (e.g. home/work commuting). Artificial neural networks may 
provide a powerful means to improve the PV yield prediction on those 
specific routes. 

The organization of the paper is as follows. Section~\ref{ReviewANN} reviews artificial feedforward neural networks, fully optimized and randomized ones, and discusses some recent theoretical results on fast training algorithms and learning guarantees in terms of consistency results and bounds for generalization errors. It draws to some extent on the expositions in \cite{Huang2004}, \cite{GabanEtAl2016} and \cite{Steland2020}.
The proposed cross-validation approach to deal with the randomness of the out-of-sample performance of randomized networks is presented in Section~\ref{Sec:CVApproach}. The application to vehicle integrated photovoltaics including a data analysis is provided in Section~\ref{Sec:Application}. 

\section{Classical Neural Networks and Extreme Learning Machines}
\label{ReviewANN}

\subsection{Hidden layer feedforward networks}
Suppose we are given input variables $ \vecz = (z_1, \dots, z_q)^\top $ and a vector $ \vecy = (y_1, \dots, y_d)^\top  $ of output variables, which are related by an assumed functional relationship disturbed by a mean zero random noise $ \bfeps $,
\[
\vecy = f( \vecz; \bftheta ) + \bfeps,
\]
where function $ f( \bullet; \bftheta ) $ is a parameterized nonlinear regression function.  Artificial neural networks form a commonly used class of regression functions and correspond to specific forms of $ f( \bullet; \bftheta ) $. In what follows, we briefly review single hidden layer feedforward networks with linear output layer  and their extension to multilayer (deep learning) networks, which have become quite popular, and discuss some recent progress in their understanding. Extreme learning machines, as introduced by \cite{Huang2004}, also called neural networks with random weights as first described by \cite{SchmidtEtAl1992}, belong to this class of artificial neural networks. 

A single hidden layer feedforward network with $q$ input nodes, $p$ hidden neurons, $d$  output neurons and activation function $g$ computes the output of the $j$th neuron of the hidden layer for an input vector $ {\vecz}_t \in \R^q $ by
\begin{equation}
	\label{ELM1}
	x_{tj} = g\left( b_j + \vecw_{j}^\top {\vecz}_t  \right), \qquad j = 1, \ldots, p.
\end{equation}
$ \vecw_j \in \R^q  $ are weights connecting the input nodes and the hidden units, and  $ b_j \in \R $ are  bias terms, $ j = 1, \ldots, p $. The output of the $j$th node of output layer is then computed as
\[
  o_{tj} = \beta_0 + \sum_{j=1}^p \beta_{j} x_{tj} = \bfbeta_j^\top \vecx_t,
\]
where $ \bfbeta_j = (\beta_0^{(j)}, \ldots, \beta_p^{(j)})^\top $ are weights and $ \vecx_t = (1, x_{t1}, \ldots, x_{tp})^\top $. There are various proposals for the activitation function. Among the most popular ones are  the sigmoid function $ g(u) = 1/(1+e^{-u}) $, the rectified linear unit (ReLU) function $ g(u) = \max(0, u)  $ or the leaky ReLU, $ g(u) = \delta x \eins( x < 0 ) + x \eins( x \ge 0 ) $, for some small $ \delta > 0 $, which allows for a small gradient when the neuron is not active. The $d$ output neurons we have $ d $ weighting vectors resulting in a $ d \times p$ parameter matrix $ \bfbeta = ( \bfbeta_1, \ldots, \bfbeta_p)^\top $ and a net output $ \veco_t = \bfbeta \vecx_t $.

Although it is well known that a single hidden layer network suffices to approximate rich classes of functions with arbitrary accuracy, see \cite[ch.~16]{GKKW2002} for an accessible treatment, multilayer (deep) learning networks are quite popular and successful for specific problems, especially in imaging. Such a 
multilayered neural network is formally defined in terms of the successive processing of the input data through $r \in \N $ hidden layers. For $r$ hidden layers with squashing functions $ g_1, \ldots, g_r $ and $ n_k $ neurons in the $k$th layer, the $j$th output of the $k$th layer is computed recursively via the equations
\[
	x_{tj}^{(1)} = g_1( b_1 + \vecx_{j}^{(1)}{}^\top \vecz_t ), \qquad j = 1, \ldots, n_1,
\]
and 
\[
	x_{tj}^{(k)} = g_k ( b_{jk} + \vecw_{j}^{(k)}{}^\top \vecx_t^{(k-1)}  ), \qquad j = 1, \ldots, n_k,  \ k = 2, \ldots, r,
\]
where $ \vecx_t^{(k)} = ( x_{t1}^{(k)}, \ldots, x_{tn_k}^{(k)} )^\top $ and  $\vecw_{j}^{(k)} $ is a $ n_k $-vector of connection weights, $k = 1, \ldots, r $.
Let $ \matW^{(k)} = ( \vecw_1^{(k)}, \ldots, \vecw_{n_k}^{(k)} )^\top \in \R^{n_k \times n_{k-1}} $ be the matrix of weights connecting the neurons of the $k$th layer with the previous layer $k-1$ resp. the input layer if $ k = 1 $, and $ \vecb_{k} = (b_{1k}, \ldots, b_{n_k,k})  \in \R^{n_k} $ the bias terms. We may write
\[
\vecx_t^{(k)} =  \vecg_k^{(\matW^{(k)}, \vecb_k)}( \vecx_t^{(k-1)} ) = g_k( \vecb_k + \matW^{(k)}  \vecx_t^{(k-1)} ), 
\]
where $ \vecg_k^{(\matW^{(k)}, \vecb_k)}( \bullet ) $ is a vector functions consisting of the $ n_k $ real-valued functions $ g_{k\ell}^{(\matW^{(k)}, \vecb_k)}( \bullet ) $, $ \ell = 1, \ldots, n_k $, and, for a real-valued function $  f $ with domain $ \R $ and a vector $ \vecu = (u_1, \ldots, u_m) \in \R^m $, $ f( \vecu) $ is defined as the vector with entries $ f( u_\ell ) $, $ \ell = 1, \ldots, m $.
To summarize, the output $ \vecx_t $ of the $r$th hidden layer for an input $ \vecz_t $ is given in terms of the composition operator $ \circ $ by
\[
	\vecx_t = \vecg_r^{(\matW^{(r)}, \vecb_r)} \circ \cdots \circ \vecg_1^{(\matW^{(1)}, \vecb_1)}( \vecz_t )
\]

\subsection{Training neural networks and extreme learning machines}

Neural networks are trained given a learning or training sample $ (\wt{\vecy}_t, \wt{\vecz}_t ), t = 1, \dots, n $ of size $n$. We denote the training in this way, in order to reserve the symbols $ \vecx_t, \vecz_t $ etc. for the validation sample addressed by the proposed statistical tools, see Section~\ref{Sec:CVApproach}.
A common approach dating back to the early days of machine learning is to minimize the least squares criterion corresponding to the quadratic loss function $ \ell( \vecu ) = \| \vecu \|_2^2 = \vecu^\top \vecu $, $ \vecu \in \R^d $,
\[
  \bftheta \mapsto L_n(\bftheta ) =  \sum_{t=1}^n \| \wt{\vecy}_t - f( \wt{\vecz}_t; \bftheta ) \|_2^2 = \sum_{t=1}^n \| \wt{\vecy}_t - \bfbeta^\top \vecx_t( \wt{\vecz}_t; \bftheta' ) \|_2^2
\]
where $ \bftheta = ( \bftheta', \bfbeta ) = (\matW_1, \ldots, \matW_r, b_1, \ldots, b_r, \bfbeta ) $ denotes the full set of parameters. This needs to be done by numerical optimization, usually a gradient-descent algorithm, where the special structure of feedforward nets allows simplified computations of the gradient by means of backpropagation, see \cite{BackProp2012} for a review. The optimizers mainly differ in how they choose the (gradient) direction and the learning rate (step size) in each step. 
Besides well known and widely used classical optimizers such as BFGS or conjugate gradient methods, stochastic gradient descent, where the algorithm cycles through the data and selects at each step the gradient evaluated at a single observations, and ADAM,  see \cite{Adam2017} and, for a proof of its local convergence, \cite{AdamConv2019}, are the most popular methods to train neural networks. 
Their efficiency in practice has certainly contributed to the success of deep learning networks. Nevertheless, the optimized artificial neural network and hence its performance in validation samples are to some extent random, since all algorithms require an initial starting value, which is chosen randomly. The best mathematical guarantee we can have is convergence at some fast rate to a local minimum, since the shape of the least squares criterion is generally known to be wigly and characterized by many local extrema, often with almost negligible curvature.

Extreme learning machines resp. neural networks with random weights make use of the following observation: If the number of neurons of the last hidden layer is equal to the number of observations, $n$, such that the output matrix of that hidden layer, $ \matX_n $, is a $ n \times n $ matrix, one can always find weights $ \bfbeta $ with  $  \matX_n \bfbeta  =  \wt{\vecY}_n  $, where $ \wt{\matY}_n = ( \wt{\vecy}_1, \ldots, \wt{\vecy}_n )^\top $  is the $ n \times d $ data matrix of the responses, {\em whatever} the values of the weights $ \bftheta' $ used to connect the remaining hidden layers among each other and the inputs with the first hidden layer. In this situation, we can perfectly explain the target values, i.e. the training data is interpolated. Here, the weights $ \bftheta' $ can also be random numbers. What happens, if $ n_T \ll n $? Then it is no longer possible to interpolate the training data and instead it makes sense to minimize the least squares criterion as a function of the weights $ \bfbeta $ of the output layer {\em given}  randomly selected weights $ \bftheta' $. This is equivalent to fitting a multiple regression model by least squares for the covariates $ \vecx_t( \wt{\vecz}_t, \bftheta' ) $ calculated for the randomly chosen $ \bftheta' $. Basically, we draw randomly a set of regressors depending on the inputs and use these regressors, which span a subspace of $ \R^n $, to explain the responses by projecting them onto that subspace. Since one no longer fully optimizes the least squares criterion with respect to all unknowns, but instead only optimizes the output layer, the computational speed up is substantial. The reason is that the latter optimization only requires to solve the normal equations $ \matX_n^\top \matX_n \bfbeta = \matY_n $ efficiently, i.e., a set of linear equations. To improve the generalization abilities it has been proposed to apply ridge regression at the output layer, also called Tikhonov regularization, which leads to the linear equations  $ (\matX_n^\top \matX_n + \lambda \matid ) \bfbeta = \matY_n $ for some regularization (ridge) parameter $ \lambda > 0 $.

\subsection{Approximation and Generalization Bounds}

Let us briefly review the general approximation abilities of such machine learners, \cite{GabanEtAl2016}. When optimizing all parameters $ \bftheta = (\vecb, \matW, \bfbeta) $ of a single hidden layer feedforward net, $ f_N(\vecz) = \sum_{j=1}^N \beta_j \vecx_j( \vecb + \matW \vecz ) $, $ \vecz \in \R^q $, it is known that the optimal achievable approximation error in the $ L_2 $-norm is independent of the dimension $q$ of the inputs and is of the order
$ O(1/N^{1/2} )$, see \cite{Barron1993}. For {\em fixed} $ \vecb, \matW $ and when optimizing only $ \bfbeta $, it has been shown that the approximation error uniformly achievable over a class of smooth functions is lower bounded by $ C / (q N^{1/q} ) $, where the constant $C$ does not depend on $N$, \cite{Barron1993}. If, however, the weights $ (\vecb, \matW) \sim \mu $ are set randomly, \cite{IgelnikPao1995} proved that the {\em expected} $ L_2 $ error is of the order $ O(1/N^{1/2} ) $. This result asserts that there {\em exists} some distribution $ \mu $ such that the expected error is $ O(1/N^{1/2} ) $. It does not contradict the worst case order of the $ L_2 $-norm of the error $ C / (q N^{1/q} ) $ for fixed weights, as it makes a statement about the average. 

Generalisation bounds for least squares estimation based on i.i.d. training samples resembling the results in \cite[ch.~3]{GKKW2002} have  been obtained by \cite{LiuEtAl2015}. In their result $ \mu $ is a uniform distribution and the mean approximation error is considered, where the expectation is taken with respect to the data distribution (as for fully-optimized feedforward networks) and with respect to $ \mu $ as well. This means, the averaged performance is studied here as well. The generalization bound is essentially also of the optimal form $ O( n^{- 2r/(2r+q) } ) $, up to a logarithmic factor, where $r$ measures the smoothness of the true regression function and $n$ is the number of samples. 

Such learning results are more informative for applications than pure approximation results, since they consider the relevant case that the artificial neural network is optimized from data using the empirical least squares criterion. But one may formulate two critiques: These results consider the framework of learning from i.i.d. samples, which is too restrictive for complex data sets. Before discussing this issue in greater detail, let us pose a second critique: The classical learning guarantees and generalization bounds address, mathematically speaking, bounds for (a functional of) the empirical generalization error which are uniform over a certain class of regression functions $f$ (i.e. networks). Such uniform bounds over a function class $ \calF $ are based on bounds for a fixed function and break the $ \sup_{f \in \calF} $ by imposing appropriate assumptions on the complexity of the class $ \calF $. Here, measures such as the Rademacher complexity, the Vapnik-Chervonenkis (VC) dimension or entropy measures provide the most satisfying and useful results, see, e.g., the monograph \cite{MohriRostaTal2018} and \cite{SH2020} for recent results for deep learners. 

But in applications, for a fixed problem and data set, the true function $ f $ is fixed, whether or not being a member of some nice class $ \calF $, and therefore the validity of learning guarantees and generalization bounds (anyway how these are defined) matters only for a single function.
It has also been criticized by \cite{N2018} that such bounds often do not explain the phenomenon that over-parameterized nets improve in terms of the test error when increasing the size of the net. The authors establish generalizations for a two-layer network, which depend on
two Frobenius matrix norms: Firstly, on the Frobenius norm of the weights of the top layer, $ \bfbeta $, and, secondly, on the Frobenius norm of $ \matW_{tr} - \matW_0 $, where $ \matW_{tr} $ denotes the trained weights and $ \matW_0 $ the randomly chosen initialization weights. The intuitive explanation of the authors is quite close to the heuristics behind extreme learning machines: If the number of hidden neurons is gets larger and finally infinity, the hidden layer provides all possible (nonlinear) features, it mainly remains to pick and combined the right ones to explain the response and tuning the weights of the hidden layers is of less importance.

Let us proceed with a discussion of the critique that the i.i.d. training framework is too restrictive for data science problems. A major issue is that many complex big data sets used in machine learning are collected over time and may also have a spatial structure.  In \cite{Steland2020} extreme learning machines and multivariate regression have been studied for a nonstationary spatial-temporal noise model having in mind data collected by moving objects (cars, drones, smartphones carried by pedestrians, etc.), which especially covers many multivariate autoregressive moving average (ARMA) time series models. Since only the weights, $ \bfbeta $, of the output layer are optimized, consistency of the least squares estimator, $ \bhbeta_n$, is of interest as well as consistency of the related prediction $ \wt{\matX}_n \bhbeta_n $ for the truth $ \wt{\matX}_n \bfbeta $. In \cite{Steland2020} it has been shown that, under quite mild regularity conditions given therein,
\[
   \| \bhbeta_n - \bfbeta \|_2^2 = O_P( p/n ), \qquad \EE \| \bhbeta_n - \bfbeta \|_2^2 = O( p/n ),
\]
where $p$ denotes the number of hidden neurons of the (last) hidden layer. This means, even for dependent noise each parameter can be estimated with the rate $ 1/\sqrt{n} $. Having in mind applications and the typical goal of prediction when fitting artificial neural networks, bounds for the sample prediction error are even more interesting. The sample mean-square prediction error (MSPE) is defined by
\[
  \wh{MSPE}_n = \frac{1}{n} \sum_{t=1}^n (\wt{\vecx}_t^\top \bfbeta - \wt{\vecx}_t^\top \bhbeta_n)^2
\]
and measures the accuracy of the predicted targets in terms of the empirical 2-vector norm with respect to the training sample. Here, and in what follows, we assume univariate targets ($d=1$).
As shown in \cite{Steland2020}, under certain conditions it holds, given the (random) weights $ \vecb, \matW $,
\[
  \wh{MSPE}_n = O_P( p/n ).
\]
For the  ridge estimator similar learning guarantees have been established generalizing and complementing result from \cite{BG2011}, \cite{Silva2014} and \cite{LiuYu2013}, which are restricted to i.i.d. sampling. Under regularity conditions given there, one can show that if the regularization parameter satisfies
\[
  \lambda_n = o_{\PP}( n / \sqrt{p} ),
\]
then the above statements on consistency of the estimated parameters of the last layer and in terms of consistency of the sample prediction error still remain true, see \cite{Steland2020}. It is worth mentioning that this result allows the regularization parameter to be random. If $ \lambda_n /n \to \lambda^0 $ for some constant $ \lambda^0 \ge 0 $, then the estimator is biased.

\section{Comparing and Cross-Validating Randomized Networks}
\label{Sec:CVApproach}

Simply training an artificial neural network to a training sample $ (\wt{\vecy}_t, \wt{\vecz}_t) $, $ t = 1, \dots, n $, should only be the first step. Comparing model specifications also taking into account additional criteria not covered by the training algorithm is generally advisable. We start with a discussion of a possible formal approach for such comparisons and evaluations. 

As argued above, any artificial neural network fitted to a training sample is random given the training sample, especially randomized nets such as extreme learning machines. As a consequence, any evaluation of the out-of-sample performance is random as well.  This implies that calculated performance measures quantifying the generalization ability are (nonnegative) random variables instead of fixed numbers. A simulation experiment discussed below demonstrates this effect.

Therefore, we propose and elaborate on a cross-validating approach using a validation sample $ (\vecy_t, \vecz_t) $, $ t = 1, \dots, n_V $, taylored for randomized networks, in order to make use of this uncertainty to improve the behaviour of the final predictions. Lastly, a method is discussed to quantify the mean sample prediction error with minimal computational costs.

\subsection{Model Comparison and Evaluation}

Suppose we are given two model specifications in terms of $ ( \vecb_1, \matW_1, \bhbeta_1 ) $ and $  ( \vecb_2, \matW_2, \bhbeta_2 ) $, where $ \vecb_i $ and $ \matW_i $ are the (random) biases and connection weights and $ \bhbeta_i $ the optimized weights of the output layer. Model comparison is often conducted by looking at the optimized values of the chosen training criterion. But the comparison or evaluation can be based on a different measure than used for training, of course, and another choice often makes sense to take into account additional objectives. Among those objectives are data fidelity, sensitivity with respect to input variables, prediction accuracy and robustness, amongst others.
Incoporating such criteria in the objective function minimized to optimize the weights of the output layer is possible, but the computational costs can increase dramatically compared with least-squares and ridge regression. 
 
Instead, one may simply select a specification (and thus a fit and prediction model) among a (small) set of candidate models, which has better behavior in terms of a selected criterion without conducting full optimizing the output layer.

Assume we have picked a criteria function $ \calC_n $ defined on $ \R^d \times \R^{n \times p} \times \R^{p+1} \to \R $ and calculate
\[
\calC_{ni} = \calC_n( \vecY_n; \wt{\matZ}_n^i, \bhbeta_i ), \qquad i = 1, 2,
\]
where $\wt{\matZ}_n^i = ( \wt{\vecz}_1^i, \dots, \wt{\vecz}_n^i  )^\top $ are the $n \times q_i $ matrices of the $n$ observations taken from $ q_i $ input variables and $ \bhbeta_i $ is the vector of optimized output layer weights, $ i = 1, 2 $. Observe that the comparison can be based on different input matrices of different dimensions. In this way, one may compare models using a different number of input variables. Especially, by putting $ \wt{\vecz}_t^{(1)} = ( \wt{z}_{t1}, \dots, \wt{z}_{tq} )^\top $ and $ \wt{\vecz}_t^{(2)} = ( 0, \dots, 0, \wt{z}_{t,q_1+1}, \dots, \wt{z}_{tq})^\top $ one can analyze whether or not the first $ q_1 $ inputs are relevant. In this case, $ \matW_2 $ is set to the last $ q-q_1 $ columns of $ \matW_1 $ and $ \vecb_2 $ to the corresponding entries of $ \vecb_1 $. A reasonable decision function is to decide in favor of model 2, if and only if the improvement expressed as a percentage is large enough, i.e. if
\[
\calC_{n2} < \calC_{n1} f 
\]
for some $ 0 < f < 1 $.

\textbf{Least squares data fidelty:} The choice
\[
\calC_n( \vecY_n; \wt{\matZ}_n^i; \bhbeta_i ) = \frac{1}{n} \sum_{t=1}^n  (Y_t - g( \vecb_i + \matW_i \wt{\vecz}_t^i)^\top \bhbeta_i )^2 
\]
corresponds to the least squares training criterion and measures the achieved data fidelity of the fit.

\textbf{Robustness:} Let $ \rho : \R \to [0, \infty) $ be a (non-decreasing, bounded, ...) function and put
\[
\calC_n( \vecY_n; \wt{\matZ}_n^i; \bhbeta_i ) = \frac{1}{n} \sum_{t=1}^n \rho( Y_t - g( \vecb_i + \matW_i \wt{\vecz}_t^i)^\top \bhbeta_i).
\] 
Here, the (loss) function $ \rho $ is used to evaluate the residuals. A common choice corresponding to robust $M$ estimation is Huber's $ \rho$-function $ \rho(u) = u^2/2 \eins( |u| \le K) + K(|u| -a/2) \eins( |u| > K ) $ for some constant $K>0$. For small $|u| $, the loss is quadratic and linear for larger values. In this way, the sensitivity to outliers is reduced.

\textbf{Mean-square prediction error:} Estimating the prediction error arising when predicting the true but unknown (optimal) mean responses by the optimized net outputs  leads to the choice
\[
\calC_n( \vecY_n; \wt{\matZ}_n^i; \bhbeta_i ) = \frac{1}{n} \sum_{t=1}^n  (\bfbeta^\top \vecx_t - \bhbeta_i^\top \wt{\vecx}_t^i )^2,
\]
where $ \wt{\vecx}_t^i = g( \vecb_i + \matW_i \wt{\vecz}_t^i) $.


The following assumption ensures that, asymptotically, the sample-based criterion function converge to constants.

\textbf{Assumption A:} $ \calC_{ni} $, $ i = 1, 2 $, converge in probability to constants $ {c}_i $, $ i = 1,2 $, i.e.
\begin{equation}
\label{DecisionRule}
\calC_{ni} \stackrel{P}{\to} {c}_i,  
\end{equation}
as $ n \to \infty $, for $ i = 1, 2 $.

Assumption A is rather weak, especially, because no rate of convergence is required. The question arises to which extent a model needs to improve upon a competitor. 

\begin{definition}
	Let us call model 2 $ (\calC_n, f) $-preferable, if the constants from Assumption~A fulfill the requirement $ c_2 < f c_1 $ for some $ f \in (0,1] $.
\end{definition}

The following result follows almost automatically from Assumption A and tells us that the rule will select the right model with probability one in large samples.

\begin{theorem}
	\label{Th1}
	Suppose that Assumption A holds true and let $ f \in (0,1] $. If model 2 is $ (\calC_n, f) $-preferable, then the decision rule (\ref{DecisionRule}) selects the correct model with probability approaching zero, i.e.
	\[
	P( \calC_{n2} > \calC_{n1} f ) \to 0, \qquad n \to \infty. 
	\]
\end{theorem}

The approach discussed above is mainly designed as an additional step when training a model from the learning sample by comparing a couple of model specifications in terms of the input variables and additional criteria such as in-sample prediction error and robustness, for a fixed choice of the (random) model parameters of the hidden layer(s). Their random choice, however, introduces uncertainty and, furthermore, the prediction accuracy should be quantified with new fresh data samples.

\subsection{A Simulation Experiment}

\begin{figure}[h]
	\includegraphics[scale=.65]{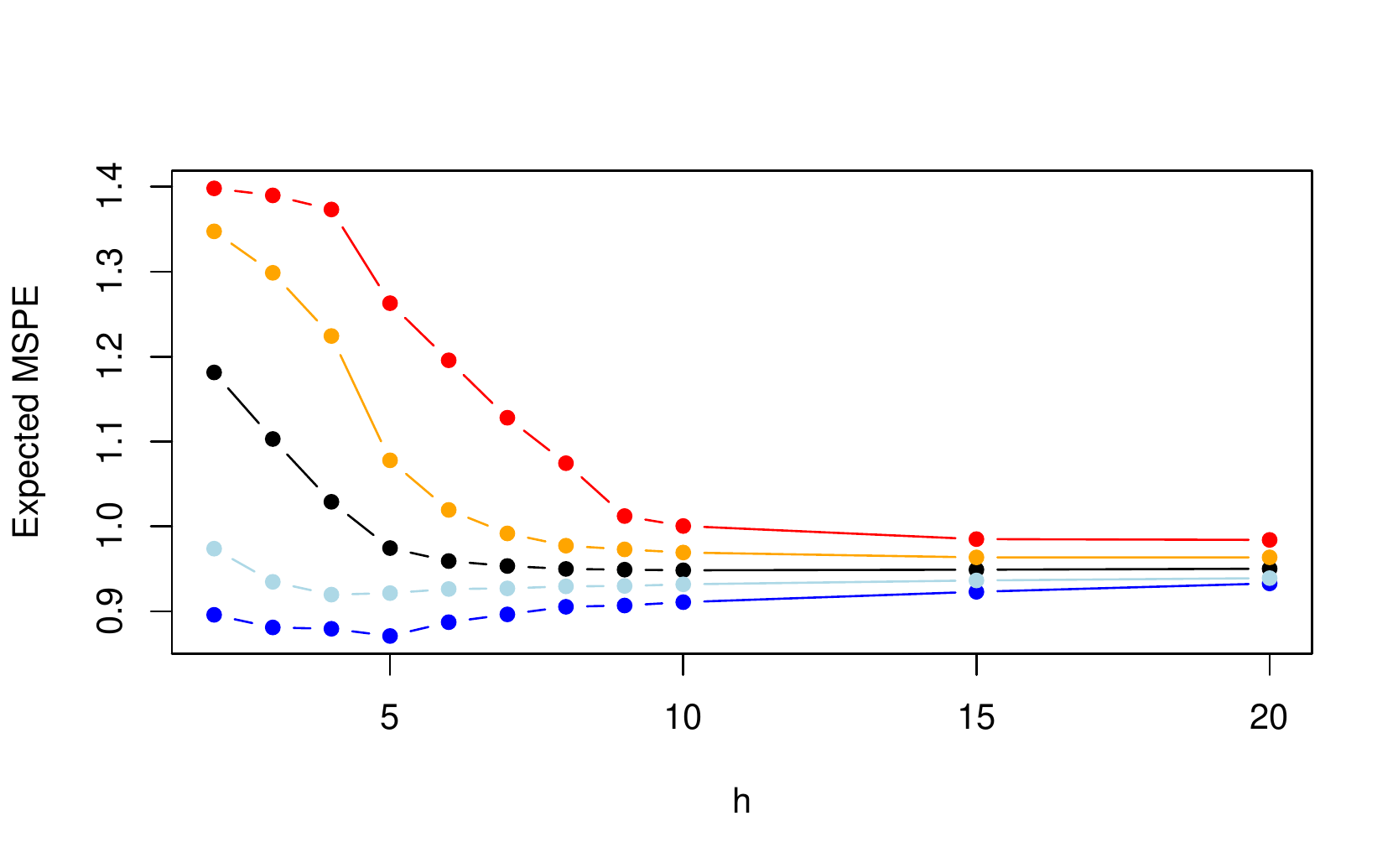}
	%
	%
	\caption{Simulated distributions of the expected sample prediction error in a validation sample of size $ 100$, for $ h = 2, \ldots, 10, 15, 20 $ hidden neurons. The curves (simulated points are joined by lines) represent from top to bottom the maximum, $ 95\% $-quantile, mean, $ 25\% $-quantile and minimum of the simulated distribution based on $1,000 $ runs.}
	\label{fig:0}       
\end{figure}

Before proceeding, let us discuss the results of a small simulation experiment conducted to illustrate the effect of randomly selecting part of the network parameters. A single hidden layer feedforward net with $ h $ neurons, $ 5 $ inputs and $ 1 $ output was examined for standard normal inputs, $ \vecz_t \sim \mathcal{N}( \vecnull, \matid ) $, and a univariate output modeled as $ y_t = \vecx_t( \vecz_t, \vecb, \matW)^\top \bfbeta_0 + \epsilon_t $, $ t = 1, \ldots, n $, where the errors $ \epsilon_t $ are i.i.d. standard normal. The training sample size was set to $ n = 1,000 $ and the validation sample size to $ n_V = 100 $. Fixing realizations of the training and test data and a randomly chosen true coefficient vector $ \bfbeta_0 $, an extreme learning machine with random weights $ \vecb, \matW $ following a uniform distribution on $ [-1,1] $ was fitted to the training sample and then evaluated in the validation sample by calculating the associated sample mean predicition error, see below for a formula. This simulation step was repeated $ 1,000 $ times to obtain for each network topology (given by $h$) an estimate of the distribution of the conditional mean predicition error. Figure~\ref{fig:0} shows characteristics of the simulated distribution as a function of the number, $h$, of hidden neurons. One can observe that the support of the distribution opens the door for picking a realization of the weights leading to superior out-of-sample performance compared to single-shot fitting of a neural network with random weights.

\subsection{Cross-Validation for Randomized Networks}

 In what follows, we assume that a validation sample $ (\vecY_i, \vecz_i ) $, $ i = 1, \ldots, n_V $, of size $ n_V $ is available for evaluation of a fitted neural network. In principle, this could be generalized to $k$-fold cross-validation scheme, but to keep the presentation simple and clean, we confine ourselves to the setting of a training of size $n$ and a validation sample of size $n_V$ as in the experiment reported above. In such a setting, one often works with ratios $ n/n_V $ around $ 80/20 $, whereas  $k$-fold cross-validation would split the available data in $k$ equal parts (folds). We elaborate on a single hidden layer network and leave the simple, notational changes for deep learning networks to the reader.

Suppose that $ Z_j = Z( \bfeta_j ) $ is a nonnegative measure for the prediction accuracy calculated for a random draw $ \bfeta_j $ of a subvector $ \bfeta $ of the full parameter $ \bftheta $ of a neural network. In case of a hidden layer net we have $ \bfeta = (\vecb,\matW) $ and $ \bftheta = (\bfeta,\bfbeta) $. In view of the randomness of $ \bfeta_j $, $ Z_j $ is a random variable attaining values in $ [0, \infty) $. For $ J $ i.i.d. draws given the training and validation samples, we obtain an i.i.d. sample $ Z_1, \ldots, Z_J $, namely the $J$ evaluations in the validation sample, when conditioning on the training and the validation samples.

The sample mean square prediction validation error for the validation data set of size $ n_V $, given in terms of the response $n_v$-vector $ \vecY_{n_V} $ and inputs $ \matZ =(\vecz_1, \ldots \vecz_{n_V} )^\top $, a $ n_V \times q $ matrix, is calculated as 
\[
\wh{MSPE}_{n_V} = \frac{1}{n_V}  \| \vecY_{n_V} - \matX_{n_V}( \vecb, \matW, \vecZ) \bhbeta_n \|_2^2 =  \frac{1}{n_V} \sum_{i=1}^{n_V} ( Y_i - g( \vecb + \matW \vecz_i )^\top \bhbeta_n )^2.
\]
The predictions $ \wh{\vecY}_{n_V} $ for the validation data set $(Y_i, \vecz_i) $, $ i = 1, \ldots, n_V $, are therefore computed using the output matrix of the hidden layer when fed with the validation inputs, i.e. using the $ n_V \times p$ output matrix
\[
\matX_{n_V} (\vecb, \matW, \vecZ) = \left[ \begin{array}{c} g( \vecb + \matW \vecz_1 )^\top \\ \vdots \\  g( \vecb + \matW \vecz_{n_V} )^\top \end{array} \right]
\]
and therefore take the form $ \wh{\vecY}_{n_V} = \matX_{n_V} (\vecb, \matW, \vecZ)  \bhbeta_n $. Here $ \bhbeta_n $ denotes the least squares estimator of the output layer calculated from the training sample $ (\wt{Y}_i, \wt{\vecz}_i ) $, $ i = 1, \ldots, n $, of size $n$, and $ \vecb, \matW $ are the random network parameters connecting the input and the hidden layer, i.e., using the $ n \times p $ output matrix of the hidden layer
\[
\wt{\matX}_{n} (\vecb, \matW, \wt{\vecZ}) = \left[ \begin{array}{c} g( \vecb + \matW \wt{\vecz}_1 )^\top \\ \vdots \\  g( \vecb + \matW \wt{\vecz}_n )^\top \end{array} \right],
\]
such that
\[
	\bhbeta_n = ( \wt{\matX}_{n} (\vecb, \matW, \wt{\vecZ})^\top \wt{\matX}_{n} (\vecb, \matW, \wt{\vecZ})  )^{-1} \wt{\matX}_{n} (\vecb, \matW, \wt{\vecZ} )^\top \wt{\vecY}_{n} 
\]

This process is now iterated $J$ times, i.e., we draw $J$ sets of random parameters $ ( \vecb(j), \matW(j) )$, $ j = 1, \ldots, J $, and calculate for each draw $ ( \vecb(j), \matW(j) ) $ the associated estimate of the sample mean prediction error in the validation sample, 
\begin{equation}
\label{DefZjs}
	Z_j = \| \vecY_{n_V}(j) - \matX_{n_V}( \vecb(j), \matW(j), \matZ ) \bhbeta_n \|_2^2.
\end{equation}
These estimates are averaged to obtain an estimator of the conditional mean $ \EE( \wh{MSPE}_{n_V} | (\wh{\vecY}_n, \wh{\matX}_n), (\vecY_{n_V}, \matX_{n_V} ) ) $, where the expectation is with respect to the distribution of the randomized network weights.  Further, we have simulated a sample of realizations of the $ \wh{MSPE}_{n_V} $ and may select the network leading to the best performance in the validation sample, i.e. take $ j^* \in \{ 1, \ldots, J \} $ with
	\[
	Z_{j^*} = \min_{1 \le j \le J} Z_j
	\]
and thus use the network with the specification $ (\vecb(j^*), \matW(j^*)) $ of the randomized parameters. This algorithm is summarized below. General results on the consistency of this approach including bounds for the estimated mean sample mean prediction error and model selection consistency are subject of ongoing research, \cite{Steland2020b}.

\vskip 0.2cm
\begin{samepage}
\begin{sf}
\textbf{Algorithm:}
	\label{Algo}
	\begin{enumerate}
		\item Draw $ (\vecb(j), \matW(j) ) \stackrel{i.i.d.}{\sim} G $, $ j = 1, \ldots, J $.
		\item For $ j = 1, \ldots, J $ do
		\item $ \qquad $ Estimate $ \bfbeta $ from the training sample using the output 
		\item[] $ \qquad $ matrix $ \wt{\matX}_n(j) = \wt{\matX}_n( \vecb(j), \matW(j), \wt{\matZ}_n ) $ giving $ \bhbeta_n(j) $.
		\item $ \qquad $ Compute the predictions of the validation sample
		\item[] $ \qquad $ $ \wh{Y}_{n_V}(j) = \matX_{n_V}(j) \bhbeta_n(j) $ with $ \matX_{n_V}(j) = \matX_{n_V}( \vecb(j), \matW(j), \matZ ) $.
		\item $ \qquad $ Compute the square prediction errors 
		\item[] $ \qquad $ $ Z_j = \frac{1}{n_V} \| \vecY_{n_V} - \wh{\vecY}_{n_V}(j) \|_2^2 $.
		\item Estimate the mean square validation prediction error by
		\item[] $ \qquad $ \[  \wh{MSPE}_{n_V,J} = \frac{1}{J} \sum_{j=1}^J Z_j \]
		\item Select the network by computing $ j^* \in \{ 1, \ldots, J \} $ with
		\[
		Z_{j^*} = \min_{1 \le j \le J} Z_j
		\]
	\end{enumerate}
\end{sf}
\end{samepage}

Observe that the above algorithm covers the case of model selection, classically understood as the  the selection of the number of hidden neurons of the net, as well as the selection of the network topology. This is so because the distribution $G$ may take into account certain toplogies, e.g., a convolutional layer which linearly processes all fixed-length subvectors of the inputs $ \vecz $ by a linear filter with randomly drawn coefficients, followed by a maxpolling layer, which is then further processed. Similarly, $G$ could be defined such that for a fully connected layer the connection weights of each neuron have  $ \ell_0 $-norm $ s_0 $, so that each neuron processes only $ s_0 $ of the outputs of the previous layer resp. of the inputs. In this way, one can try (randomly) different network topologies in a systematic way.

\subsection{An Uncertainty Interval for the Mean Sample Prediction Error with Minimal Computational Costs}

In order to deal with the uncertainty of the sample MSPE and to minimize the required computational costs, one can calculate a fixed-width confidence interval for the expected sample MSPE in the validation sample,  $ \mu = \EE_{(\vecb,\matW)}\left( \wh{MSPE}_{n_V}  \right) $, corresponding to the black points in Figure~\ref{fig:0}. For a fixed uncertainty $d>0$, specified in advance as the half-length of an interval aroung the estimator  $ \wh{MSPE}_{n_V,J} $, one wants to determine $J$ from data, such that  the resulting interval has confidence $ 1-\alpha $, $ \alpha >0 $ small.  This means, we want to determine the smallest $ J $, such that the fixed-width interval 
\[
  \left[ \wh{MSPE}_{n_V,J} - d, \wh{MSPE}_{n_V,J}  + d   \right]
\]
has coverage probability $ 1 - \alpha $. The problem to construct fixed-width uncertainty intervals has been studied for general parameters by \cite{ChangSteland2020} for the classical asymptotic regime $ d \to 0 $ as well as the novel high-confidence regime $ 1-\alpha \to 1 $. A solution, $ \wh{J}_{opt} $, which is consistent for the theoretically optimal solution, and first as well as second order efficient, is as follows: One fixes a minimal number of draws, $ \bar{J}_0 $, and calculates 
\[
  J_0 = \max \left\{  \bar{J}_0, \left\lfloor  \frac{\Phi^{-1}( 1-\alpha/2) \wh{\sigma} }{d}  \right\rfloor +1  \right\}.
\]
Here $ \Phi^{-1} $ is the quantile function of the standard normal distribution function.
 $ \widehat{\sigma} $ is the sample standard deviation of $ Z_j $'s of a small number of initial runs, which can be as small as $3$ according to the simulation studies in \cite{ChangSteland2020}. Next, perform $ J_0 $ simulation runs and calculate $ \wh{\sigma}_{J_0}^2 = \frac{1}{J_0} \sum_{j=1}^{J_0} (Z_j - \overline{Z} )^2 $. Lastly, 
 one calculates the final number of runs given by
\[
  \wh{J}_{opt} = \max\left\{  J_0, \left \lfloor \frac{ \wh{\sigma}_{J_0}^2 \Phi^{-1}(1-\alpha/2)^2 }{d^2}  \right\rfloor  \right\}.
\]
If $  \wh{J}_{opt} > J_0 $, one conducts the required additional $ \wh{J}_{opt} - J_0 $ draws of the random parameters of the neural net, determines the associated sample prediction errors $ Z_1, \ldots, Z_{J_{\wh{J}_{opt} }} $ according to (\ref{DefZjs}), and eventually computes the interval $ \overline{Z}_{\wh{J}_{opt} } \pm d $.

\section{Application to Vehicle Integrated Photovoltaics and Data Analysis}
\label{Sec:Application}

An interesting specific problem arising in VIPV is the prediction of the yield due to the integrated solar panels. Compared to panels mounted at the rooftop of a truck or car, panels mounted at the sides pose additional problems, since  their energy yield depends on the orientation of the vehicle. The question arises to which extent one can predict their contribution to the total yield by the irradiance measured at the rooftop.  The basic idea to explain the  measurements of a sensor (or PV module) facing left (or right) in terms of a sensor facing up is that it is easier to derive, in advance, expected irradiance maps for horizontally aligned sensors. To a planned route one can then assign an expected irradiance trajectory for the sensor facing up. A prediction model then allows us to forecast the contribution of further sensors. In this way, one can predict the VIPV yield. 

\subsection{Vehicle Mounted Data Logger}
Several sensors and a data logger were mounted on a vehicle. The sensors 
include 4 irradiance sensors, one acoustic wind sensor, a 
Global Positioning System (GPS), and a magnetometer. 

The sensors are specifically chosen to provide relevant data for VIPV 
yield. For this purpose we obviously want to monitor the irradiance. 
The irradiance depends strongly on the orientation of the PV module. 
Thus, we use 4 irradiance sensors facing in different directions (top, 
left, right and back), and we log the vehicle orientation. While the 
vehicle is moving we can use GPS data to provide a good indicator for 
the vehicle orientation (assuming the vehicle is moving in forward 
direction). However, as vehicles are also often parked, we in addition 
use a magnetic sensor to provide information on the vehicle orientation. 

Another important factor for yield is the module temperature, as PV 
modules are less efficient at higher temperatures. The module 
temperature itself depends on several environmental factors; wind, 
ambient temperature, and irradiance. In \cite{Kuehnel2017VIPV} it was 
shown that the head wind from driving provides a significant positive 
impact on PV yield as the additional wind cools the PV modules. 

The data logger was developed around a Raspberry Pi single board 
computer. The Raspberry Pi is equipped with a GPS module and a 
magnetometer. Note that the magnetometer is used as GPS only provides 
information on the orientation of the vehicle while the vehicle is 
moving (assuming the vehicle is mover forward). However, most vehicle 
spend a large amount of time parked. The remaining wind and irradiance 
sensors are connected with two RS485 interfaces, one for the wind 
sensor and one for the four irradiance sensors. The logged sensor data 
is written to a USB thumb drive. The setup is powered from the 12 V car 
battery and is enclosed in a weather proof box mounted on a rooftop 
rack.

For the irradiance sensors we used four calibrated silicon sensors from 
Ingenieurb\"uro Mencke \& Tegtmeyer GmbH of type SiRS485TC-T-MB. As the 
sensors are silicon reference cells the measured irradiance is of 
particular relevance for PV applications as the spectral range of the 
sensors matched that of typical PV modules. The four irradiance sensors 
are mounted on the same rooftop rack, facing up, left, right and 
backwards. 

The wind sensor is an FT205 acoustic wind sensor from FT Technologies. 
The sensor measures both wind speed and direction (2D). The sensor also 
reports the acoustic air temperature, i.e. the air temperature derived 
from the temperature dependent speed of sound in air.

The data used in this paper was collected during several test drives of 
the system. We plan to use several car mounted data logging systems in 
the coming years on several cars with different use profiles.

\subsection{Data Analysis}

As a preliminary study, we analyzed a small pilot sample collected during three test drives. In view of the limited data available for this analysis, we can only get a first impression whether the information describing the position of the car, namely where it is located and in which direction it drives, can be exploited to predict measurements of a sensor facing left, right or backwards from measurements from the sensor facing up. 

The available data was split in a training sample with $ n = 3,472 $ data points and a test sample with $ 3,669 $ observations. The validation sample was selected as observations $ 1,000-2,250 $ from the test sample, since such PV data is highly heterogenous, as irradiance differs substantially depending on the time of day and weather. For the present data set, the first part of the test sample was inappropriate. 

In our nonlinear model it is assume that the $t$th voltage measurement of the sensor facing left, $ s_{2t} $,  is related to the sensor facing up, $ s_{1t} $, via the equation
\[
  s_{2t} = s_{1t}(1+f( a_t, x_t, y_t ) ), \qquad t = 1, \dots, n.
\]
Here $ a_t $ denotes the angle (direction) of the car and $ (x_t, y_t) $ is the car's location at time $t$, expressed in terms of geographical coordinates (langitude and latitude). $f$ is an unknown (nonlinear) function. A baseline (null) model would be to assume that $f$ is equal to some constant value $ f_0 $. It is, however, clear that under idealized noiseless conditions, $f$ is a function of angle and geographical location. For example, at a certain location the car's side but not the roof may be shadowed by a building. Of course, a more refined model needs to take into account time of day and season, but estimating such models requires sufficiently big data set over much longer time span than available for the present illustrative data analysis.

The function $f$ can be modeled and estimated by a nonlinear regression approach,
\[
  y_{t} = f( a_t, x_t, y_t ) + \epsilon_t, \qquad t = 1, \ldots, n,
\] 
for mean zero random noise terms $ \epsilon_t $, using the targets (responses)
\[
  y_{t} = \frac{s_{2t}- s_{1t}}{s_{1t}}
\]
and the input variables (regressors) $ \vecz_t = ( \alpha_t, x_t, y_t ) $. Having a prediction $ \wh{y}_t $ the corresponding forecast of $ s_{2t} $ is then calculated as $ \wh{s}_{2t} = s_{1t} ( 1 + \wh{y}_t ) $.

We compared two model specifications. Firstly, a linear specification, i.e., a classical multiple linear regression model, given by
\[
  y_t = b_0 + b_1 a_t + b_2 x_t + b_3 y_t + \epsilon_t,
\]
for regression coefficients $ b_0, \ldots, b_3 \in \R $. The second model is a single hidden layer feedforward network with $ p = 4 $ hidden units and a logistic squasher $ 1/(1+\exp(-x)) $,
\[ 
  y_t = f( \vecz_t; \bftheta ) + \epsilon_t = \vecx_t( \vecz_t; \bfeta )^\top \bfbeta + \epsilon_t,
\]
where $ \bftheta = (\bfeta, \bfbeta) $ with $ \bfeta= ( \vecb, \matW ) $, and $ \bfbeta \in \R^p$ represents the weights of the linear output layer. It is worth mentioning that fitting successfully neural networks requires to norm the input variables to the interval $ [-1,1] $. The neural net was trained as an extreme learning machine using ridge regression with ridge regularization parameter $ \lambda = 0.2 $ and random weights $ \bfeta $ with i.i.d. entries following a uniform distribution on the interval $ [-1,1] $.  In order to get more robust results, the most extreme $ 5\% $ of the observations of the training sample were omitted. Following the proposed cross-validation method, a realization of $ \eta^* = (\vecb^*, \matW^*) $ was chosen which yields the best prediction accuracy in the validation sample.

Table~\ref{tab:1} provides the sample mean prediction error in the validation sample for all three prediction methods, the baseline null model, multiple linear regression and artificial neural network. In addition, for each model the statistic $ \wh{MSPE}_{n_V} = \frac{1}{n_V} \sum_{i=1}^n \wh{e}_i $, where $ \wh{e}_i $ denotes the prediction error for the $i$th datapoint, e.g., 
$ \wh{e}_i = Y_i - \vecx_i( \vecb^*, \matW^*, \vecz_i)^\top \bhbeta_n $ for the neural network,
was decomposed by computing the components
\begin{align*}
\wh{MSPE}_{n_V,0.1} &= \frac{1}{n_V} \sum_{i=1 \atop | \wh{e}_i | \le q_{0.1}  }^{n_V} \wh{e}_i^2, \\
\wh{MSPE}_{n_V,0.1:0.9} &= \frac{1}{n_V} \sum_{i=1 \atop q_{0.1} < | \wh{e}_i | < q_{0.9}  }^{n_V} \wh{e}_i^2, \\
\wh{MSPE}_{n_V,0.9} & = \frac{1}{n_V} \sum_{i=1 \atop | \wh{e}_i | > q_{0.9}  }^{n_V} \wh{e}_i^2, 
\end{align*}
where $ q_p $ denotes the $p$-quantile of the empirical distribution of the prediction errors $  \wh{e}_i  $, $ i = 1, \ldots, n_V $. In this way, one can analyze how well a method works in the tails compared with the central $ 90\% $ of the data.  $ \wh{MSPE}_{n_V,0.1} $ measures the prediction error when the method overestimates and $  \wh{MSPE}_{n_V,0.9} $ if it underestimates. One can observe that the neural net predictions surprisingly well in the central part and also when it overestimates, but the predicition errors are large when it underestimates.

\begin{table}[!t]
	\caption{Prediction accuracy in the validation sample.}
	\label{tab:1}       
	%
	%
	\begin{tabular}{p{2.7cm}p{2.1cm}p{2.1cm}p{2.1cm}p{2.1cm}}
		\hline\noalign{\smallskip}
		Method  & $\wh{MSPE}_{n_V}$ & $ \wh{MSPE}_{n_V,0.1} $ & $ \wh{MSPE}_{n_V,0.1:0.9} $ & $ \wh{MSPE}_{n_V,0.9} $  \\
		\noalign{\smallskip}\hline\noalign{\smallskip} 
		Null model & 100,555.8  & 11,635.59  &  6,908.67  &  82,011.5 \\
		Linear regression & 67,967.4 & 12,740.26  &  5,547.23  &  49,679.9  \\
		ELM neural network & 74,502.3 &  4,065.44  &  2,133.83 & 68,303.0  \\
		\noalign{\smallskip}\hline\noalign{\smallskip}
	\end{tabular}
\end{table}

\begin{figure}[h]
	\includegraphics[scale=.65]{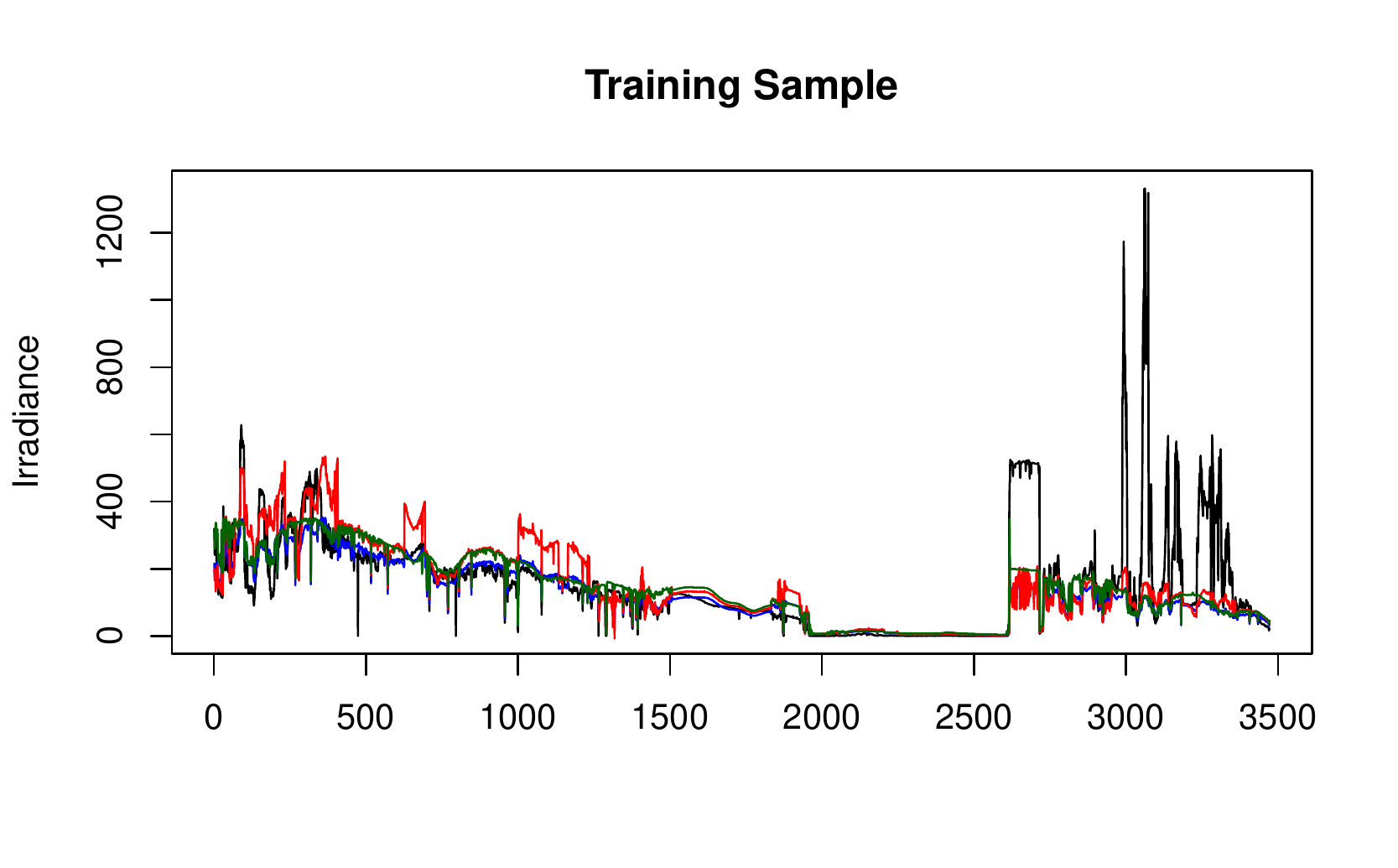}
	%
	%
	\caption{Observed irradiance at sensor 2 and predictions for the training sample: Null model (green), linear regression (red), extreme learning machine (blue).}
	\label{fig:1}       
\end{figure}

Figure~\ref{fig:1} shows the predictions of the three prediction methods in the training sample, whereas Figure~\ref{fig:2} depicts the results for the validation and test sample. 
The predictions of the nonlinear neural network are in most cases closer to the observed data points, except for some extreme measurements, which are not nicely captured by the neural net.  In Figure~\ref{fig:2} the cumulated measurements and their predictions, respectively, are plotted. Since the sensors provide data sampled at a fixed sampling rate without gaps, the cumulated values can be regarded as proxies for the (total) energy yield. Because the neural network is not able to capture some extremes, it underestimates the yield. 

However, the data set used in this pilot study is too small to draw conclusions, especially about the question to which extent artificial neural networks outperfrom linear methods for the problem of interest. It is also not clear whether the observed properties of the prediction errors are artifacts or will still be present when larger data sets are analyzed.

\begin{figure}[h]
	\includegraphics[scale=.65]{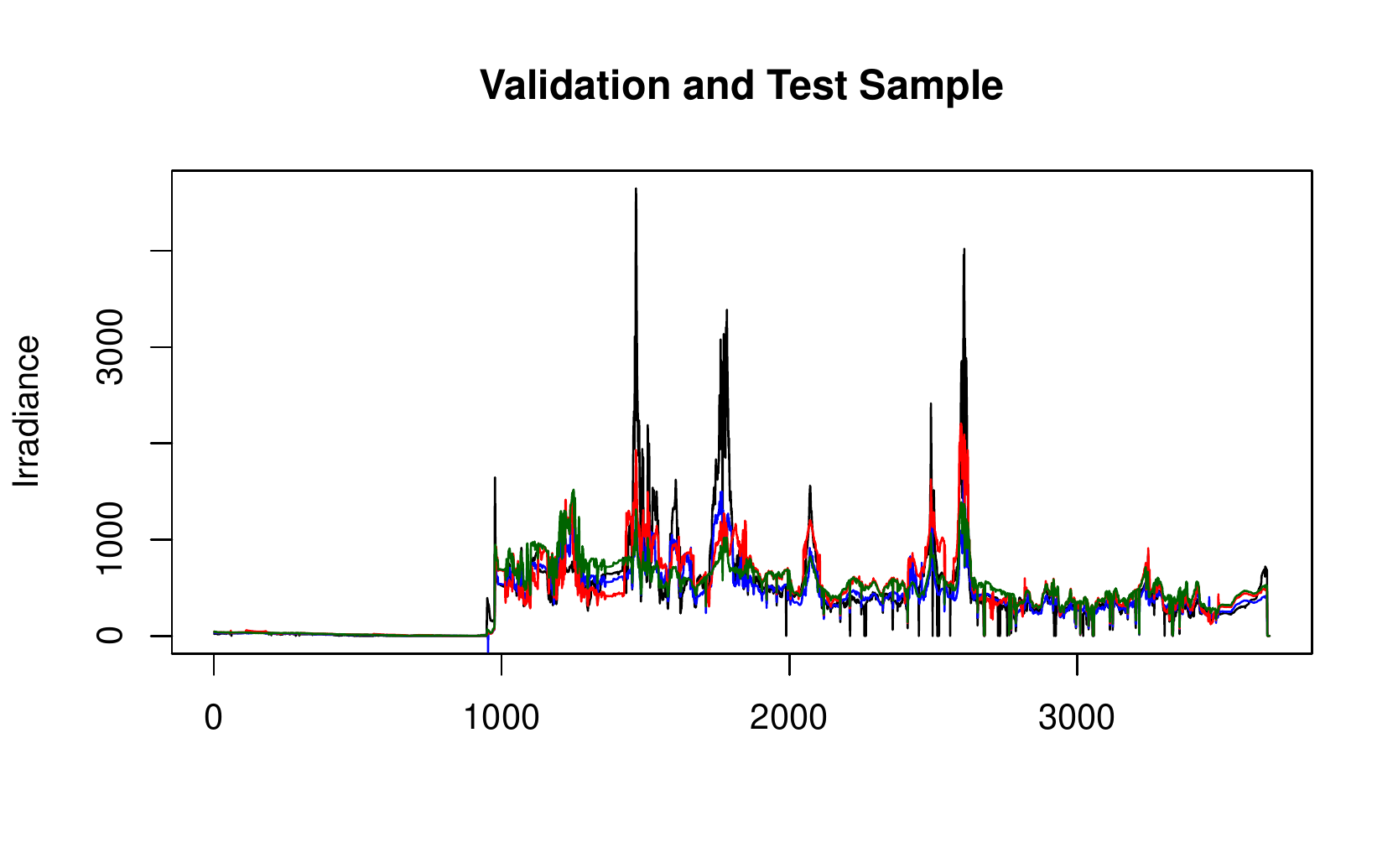}
	%
	%
	\caption{Observed irradiance and predictions for the validation and test sample: The neural net yields better predictions in most cases, but underestimates several extrema.}
	\label{fig:2}       
\end{figure}

\begin{figure}[h]
	\includegraphics[scale=.65]{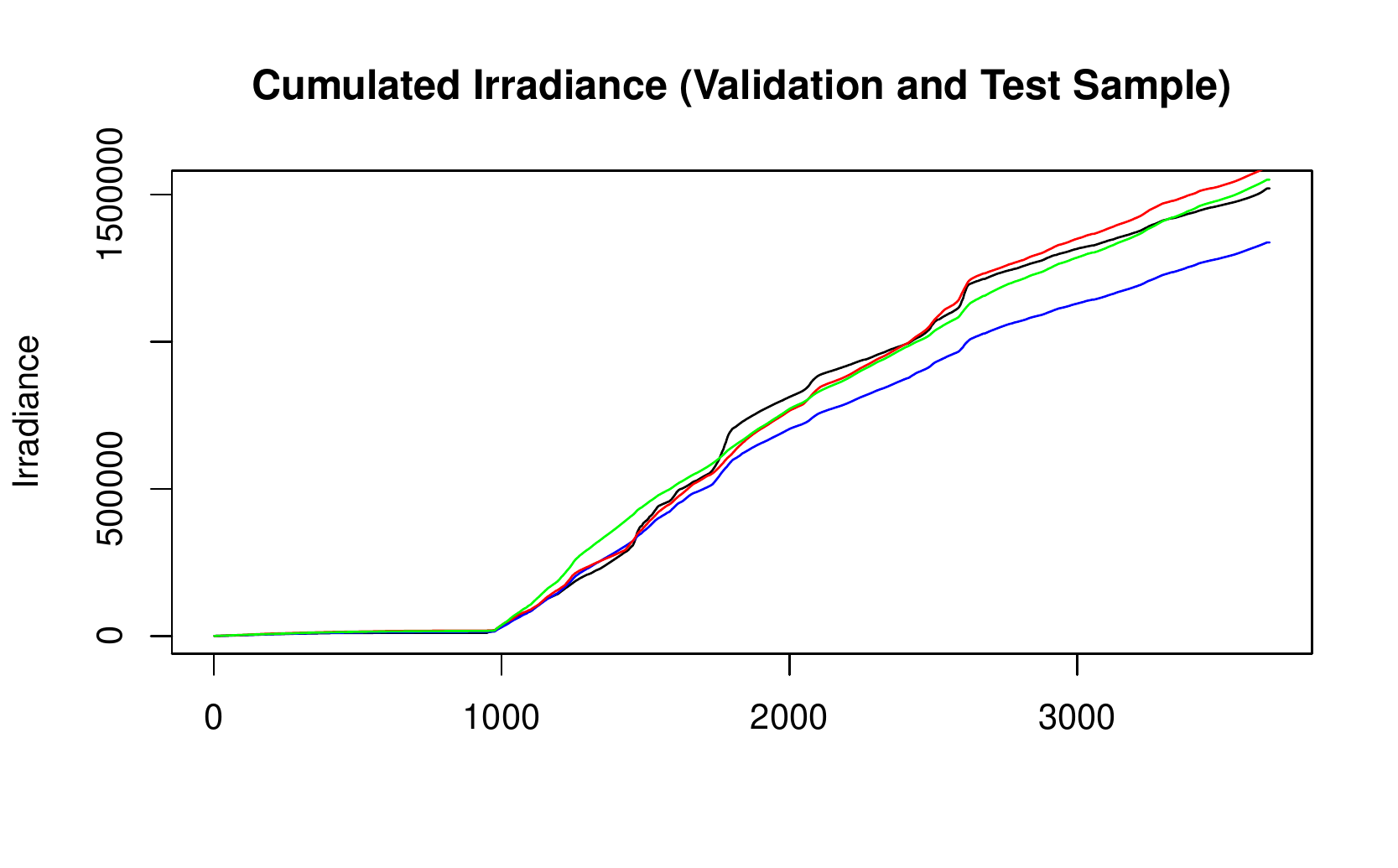}
	%
	%
	\caption{Observed cumulated  measurements  and cumulated predictions for the validation and test sample: The neural net underestimates power generation in the test sample as several extrema are not properly predicted.}
	\label{fig:2}       
\end{figure}

\section*{Appendix: Proof of Theorem~\ref{Th1}}

If model 2 is $ (\calC_n, f) $-preferable, then the probability of a false decision is given by
\[
  P( \calC_{n2} > \calC_{n1} f ) 
  = P( \calC_{n2} - \delta_2 > \calC_{n1} f - c_1 f - \delta_2 + c_1 f )  
\]
Consequently,
\begin{align*}
  P( \calC_{n2} > \calC_{n1} f ) 
  & = P( [\calC_{n2} - c_2] + [\calC_{n1}-c_1]f > c_1f - c_2 ) \\
  & \le P( [\calC_{n2} - c_2] > (c_1f-c_2)/2 ) + P(  [\calC_{n1}-c_1] > (c_1f - c_2)/(2f) ) \\
  & \to 0,
\end{align*}
as $ n \to \infty $, since $ c_1 f - c_2 > 0 $. From these simple bounds it is clear that a convergence rate for the criterion automatically yields a convergence rate for the error probability to select the wrong model.

\section*{Acknowledgement} This work has been financially supported by the German Federal Ministry for Economic Affairs and Energy (Bundesministerium f\"ur Wirtschaft und Energie, BMWi), within the project 
'Street -- Einsatz von hocheffizienten Solarzellen in elektrisch betriebenen Nutzfahrzeugen', grant no. 0324275A. The authors are responsible for the content.	They  gratefully acknowledge preparatory work of M.Sc. Nils Br\"ugge.

\end{document}